\documentclass[11pt,a4paper]{article}
\usepackage[hyperref]{tacl2018v2}

\usepackage[T1]{fontenc}
\usepackage{times}
\usepackage{plex-mono}

\usepackage{latexsym}
\usepackage{amsmath,amssymb}
\usepackage{dsfont}

\usepackage{algorithm}
\usepackage{algpseudocode}

\usepackage{array,etoolbox}
\preto\tabular{\setcounter{magicrownumbers}{0}}
\newcounter{magicrownumbers}

\usepackage{tikz}
\usetikzlibrary{arrows,matrix,positioning,fit,calc,shapes,decorations.pathreplacing,svg.path,shadows,patterns}
\pgfdeclarelayer{background}
\pgfdeclarelayer{backgroundoverlay}
\pgfsetlayers{background,backgroundoverlay,main}
\usepackage{pgfplots}
\pgfplotsset{compat=1.13}
\usepackage{adjustbox}
\usepackage{ocr}

\usepackage{dirtytalk}
\usepackage{microtype}

\usepackage{booktabs}
\usepackage{arydshln}
\usepackage{multirow}
\usepackage{array}
\newcolumntype{H}{>{\setbox0=\hbox\bgroup}c<{\egroup}@{}}

\usepackage{bm}

\newcommand{\tokdetok}[0]{\textsc{XRayEmb}}
\newcommand{\tok}[0]{\textsc{XR-Enc}}
\newcommand{\detok}[0]{\textsc{XR-Dec}}
\newcommand{\tdloop}[0]{E$\rightarrow$D}
\newcommand{\dtloop}[0]{D$\rightarrow$E}
\newcommand{\marco}[0]{\textsc{MarcoQA}}

\newcommand{\llm}[0]{LPLM}
\newcommand{\wpc}[0]{WordPiece}
\newcommand{\ppt}[0]{\textsc{2pt}}
\newcommand{\mmod}[0]{\textsc{$M$}}
\newcommand{\gen}[0]{\textsc{Pred}}
\DeclareMathOperator*{\agg}{agg}

\renewcommand{\vec}[1]{\bm{#1}}

\newcommand{\vE}{\ensuremath \vec{E}}
\newcommand{\vI}{\ensuremath \vec{I}}
\newcommand{\vc}{\ensuremath \vec{c}}
\newcommand{\vh}{\ensuremath \vec{h}}

\newcommand{\ve}{\ensuremath \vec{e}}

\taclpubformattrue 

\title{Learning to Look Inside: Augmenting Token-Based Encoders with Character-Level Information}

\author{Yuval Pinter\thanks{Work done as an intern at Bloomberg LP and as a Bloomberg PhD Data Science Fellow.} \\
  School of Interactive Computing \\
  Georgia Institute of Technology \\
  Atlanta, GA, USA \\
  \texttt{uvp@gatech.edu} \\\And
  Amanda Stent \\
  Bloomberg \\
  New York, NY, USA \\
  \texttt{astent@bloomberg.net} \\\AND 
  Mark Dredze \\
  Bloomberg \\
  Department of Computer Science \\
  Johns Hopkins University \\
  \texttt{mdredze@cs.jhu.edu} \\\And
  Jacob Eisenstein \\
  Google Research \\
  \texttt{jeisenstein@google.com} \\}

\date{}

\begin{document}
\maketitle
\begin{abstract}
    Commonly-used transformer language models depend on a tokenization schema which sets an unchangeable subword vocabulary prior to pre-training, destined to be applied to all downstream tasks regardless of domain shift, novel word formations, or other sources of vocabulary mismatch.
    Recent work has shown that ``token-free'' models can be trained directly on characters or bytes, but training these models from scratch requires substantial computational resources, and this implies discarding the many domain-specific models that were trained on tokens. In this paper, we present \tokdetok{}, a method for retrofitting existing token-based models with character-level information.  
    \tokdetok{} is composed of a character-level ``encoder'' that computes vector representations of character sequences, and a generative component that decodes from the internal representation to a character sequence.
    We show that incorporating \tokdetok{}'s learned vectors into sequences of pre-trained token embeddings helps performance on both autoregressive and masked pre-trained transformer architectures and on both sequence-level and sequence tagging tasks, particularly on non-standard English text.
\end{abstract}

\section{Introduction}
\label{sec:intro}

The use of subword representations in NLP applications has become the default choice in recent years, most notably within the framework of large pre-trained language models (\llm{}s) where the LM objective requires a vocabulary that is \textit{computationally feasible} for a softmax operation, typically interpreted as no larger than $10^{5}$ types, while also \textit{exhaustive}, such that it allows production of arbitrary space-delimited tokens in order to mitigate the out-of-vocabulary (OOV) problem.
Informative subword vocabularies are pre-computed by methods such as byte-pair encoding~\cite[BPE;][]{sennrich-etal-2016-neural} and its variant \wpc{} used in BERT~\cite{devlin-etal-2019-bert}, or Unigram LM~\cite{kudo-2018-subword}.
Such vocabularies normally make use of explicit signalling for either word-medial or word-boundary tokens, so that recompiling the original space-delimited words is a deterministic process.\footnote{In this paper, we refer to space-delimited sections of text as \say{words} even when they are not linguistic words per se, in order to distinguish them from the model-centric \say{tokens} which are represented and operated over by \llm{}s.}

\begin{table*}
    \centering
    \small
    \begin{tabular}{c} \toprule
        \#simplyleopard \#4thofjuly \#showmeyourmumu @ Town of Breckenridge, Colorado \\ \midrule
        \# simply \#\#le \#\#opa \#\#rd \# 4th \#\#of \#\#ju \#\#ly \# show \#\#me \#\#you \#\#rm \#\#um \#\#u \\
        @ town of br \#\#eck \#\#en \#\#ridge , colorado \\
        \bottomrule
    \end{tabular}
    \caption{A tweet (top) and its \wpc{} tokenization  (\textsc{bert-base-uncased}, bottom).} 
    \label{tab:example}
\end{table*}

However, processes for constructing subword vocabularies have high variance: perturbing the dataset used to train them leads to significant changes in the resulting vocabulary.
In a preliminary experiment, we sampled two sets of 1 million sentences from the same Wikipedia dump and trained a Unigram LM model of 32,000 tokens on both.
The resulting vocabularies presented a discrepancy of 27\% (8,620 unshared tokens).
Similarly, \newcite{lazaridou2021pitfalls} found that collecting corpus data from different timestamps of the same source shifts the vocabulary towards terms used more frequently over different times.
Nevertheless, this creation procedure is irreversible: once a subword vocabulary has been set, the \llm{} is tied down to it regardless of data encountered in subsequent LM training and fine-tuning.

\begin{figure*}[ht]
    \centering
    \begin{tikzpicture}
    
        \tikzstyle{green_circle} = [shape=circle, minimum size=0.2cm, circular drop shadow, text=black, very thick, draw=black!55, top color=white,bottom color=green!80, text width=0.2cm, align=center]
        \tikzstyle{red_circle} = [shape=circle, minimum size=0.2cm, circular drop shadow, text=black, very thick, draw=black!55, top color=white,bottom color=red!80, text width=0.2cm, align=center]
        \tikzstyle{embs} = [shape = rectangle, rounded corners]
        
        \tikzset{>=stealth'}
        
        \node at (-8.5,-5) {$X$};
        \draw [thin, double, rounded corners] (-7,-4.5) rectangle (6.5,-5.5);
        \node at (-6,-5) {My};
        \node (v3) at (-3.5,-5) {hovercraft};
        \node (v1) at (-0.5,-5) {is};
        \node at (1,-5) {full};
        \node at (2.5,-5) {of};
        \node at (4,-5) {eels};
        \node at (5.5,-5) {.};
        
        \node at (-8.5,-3.75) {$\tau$};
        \draw [thin, double, rounded corners] (-6.5,-3.5) rectangle (-5.5,-4);
        \draw [fill=white,thin, double, rounded corners] (-5,-3.5) rectangle (-3.55,-4);
        \draw [thin, double, rounded corners] (-3.45,-3.5) rectangle (-2,-4) node (v2) {};
        \draw [fill=gray!15, thin, double, rounded corners] (-1.25,-3.5) rectangle (0.25,-4);
        \draw [thin, double, rounded corners] (0.5,-3.5) rectangle (1.5,-4);
        \draw [fill=gray!15, thin, double, rounded corners] (1.75,-3.5) rectangle (3.25,-4);
        \draw [thin, double, rounded corners] (3.5,-3.5) rectangle (4.5,-4);
        \draw [thin, double, rounded corners] (5,-3.5) rectangle (6,-4);
        \node (t1) at (-6,-3.82) {my};
        \node (t2) at (-3.5,-3.25) {};
        \node (t21) at (-4.25,-3.75) {hover};
        \node (t22) at (-2.75,-3.75) {\#\#craft};
        \node (t3) at (-0.5,-3.78) {\small [MASK]};
        \node (t4) at (1,-3.75) {full};
        \node (t5) at (2.5,-3.78) {\small [MASK]};
        \node (t6) at (4,-3.75) {eels};
        \node (t7) at (5.5,-3.75) {.};
        
        \node at (-8.5,-2.5) {$\pi^t$};
        \node (p1) [green_circle] at (-6,-2.5) {\small \checkmark};
        \node (p2) [red_circle] at (-3.5,-2.5) {\small $\times$};
        \node (p3) [green_circle] at (-0.5,-2.5) {\small \checkmark};
        \node (p4) [green_circle] at (1,-2.5) {\small \checkmark};
        \node (p5) [green_circle] at (2.5,-2.5) {\small \checkmark};
        \node (p6) [green_circle] at (4,-2.5) {\small \checkmark};
        \node (p7) [green_circle] at (5.5,-2.5) {\small \checkmark};
        
        \draw[->]  (t1) edge (p1.south);
        \draw (t21) -- (t2) -- (t22);
        \draw[->]  (t2) edge (p2.south);
        \draw[->]  (t3) edge (p3.south);
        \draw[->]  (t4) edge (p4.south);
        \draw[->]  (t5) edge (p5.south);
        \draw[->]  (t6) edge (p6.south);
        \draw[->]  (t7) edge (p7.south);
        
        \node at (-8.5,-1.5) {$\mathbf{E}$};
        \draw [thin, rounded corners, fill=yellow!25] (-6.65,-1.3) rectangle (-5.35,-1.7);
              \node [matrix] (e1) at (-6,-1.5) {
              \draw [fill=orange!55] circle (1mm); & \draw [fill=orange!55] circle (1mm); & \draw [fill=orange!55] circle (1mm); & \draw [fill=orange!55] circle (1mm); & \draw [fill=orange!55] circle (1mm); \\
              };
        \draw [thin, rounded corners, fill=yellow!25] (-1.15,-1.3) rectangle (0.15,-1.7);
              \node [matrix] (e3) at (-0.5,-1.5) {
              \draw [fill=orange!55] circle (1mm); & \draw [fill=orange!55] circle (1mm); & \draw [fill=orange!55] circle (1mm); & \draw [fill=orange!55] circle (1mm); & \draw [fill=orange!55] circle (1mm); \\
              };
        \draw [thin, rounded corners, fill=yellow!25] (0.35,-1.3) rectangle (1.65,-1.7);
              \node [matrix] (e4) at (1,-1.5) {
              \draw [fill=orange!55] circle (1mm); & \draw [fill=orange!55] circle (1mm); & \draw [fill=orange!55] circle (1mm); & \draw [fill=orange!55] circle (1mm); & \draw [fill=orange!55] circle (1mm); \\
              };
        \draw [thin, rounded corners, fill=yellow!25] (1.85,-1.3) rectangle (3.15,-1.7);
              \node [matrix] (e5) at (2.5,-1.5) {
              \draw [fill=orange!55] circle (1mm); & \draw [fill=orange!55] circle (1mm); & \draw [fill=orange!55] circle (1mm); & \draw [fill=orange!55] circle (1mm); & \draw [fill=orange!55] circle (1mm); \\
              };
        \draw [thin, rounded corners, fill=yellow!25] (3.35,-1.3) rectangle (4.65,-1.7);
              \node [matrix] (e6) at (4,-1.5) {
              \draw [fill=orange!55] circle (1mm); & \draw [fill=orange!55] circle (1mm); & \draw [fill=orange!55] circle (1mm); & \draw [fill=orange!55] circle (1mm); & \draw [fill=orange!55] circle (1mm); \\
              };
        \draw [thin, rounded corners, fill=yellow!25] (4.85,-1.3) rectangle (6.15,-1.7);
              \node [matrix] (e7) at (5.5,-1.5) {
              \draw [fill=orange!55] circle (1mm); & \draw [fill=orange!55] circle (1mm); & \draw [fill=orange!55] circle (1mm); & \draw [fill=orange!55] circle (1mm); & \draw [fill=orange!55] circle (1mm); \\
              };
        \draw[->]  (p1) edge (e1.south);
        \draw[->]  (p3) edge (e3.south);
        \draw[->]  (p4) edge (e4.south);
        \draw[->]  (p5) edge (e5.south);
        \draw[->]  (p6) edge (e6.south);
        \draw[->]  (p7) edge (e7.south);
            
        \node at (-8.5,-0.5) {\tok};
        \draw [thin, double, rounded corners] (-5.2,-0.75) rectangle (-1.8,-1.2);
        \node (ph2) at (-3.5,-1) {\tiny {\texttt{h o v e r c r a f t /w}}};
        \draw [thin, rounded corners, fill=blue!25] (-4.15,0) rectangle (-2.85,-0.4);
              \node [matrix] (e2) at (-3.5,-0.2) {
              \draw [fill=blue!55] circle (1mm); & \draw [fill=blue!55] circle (1mm); & \draw [fill=blue!55] circle (1mm); & \draw [fill=blue!55] circle (1mm); & \draw [fill=blue!55] circle (1mm); \\
              };
        \begin{pgfonlayer}{background}
              \draw[bend left, ->] (v3) edge (ph2);
        \end{pgfonlayer}
        \draw[->]  (ph2) edge (e2.south);
        
        \node at (-8.5,1.5) {\mmod{}};
        \draw [fill=gray!25, thin, double, rounded corners] (-7,2.5) rectangle (6.5,0.5);
        \draw[->]  (e1) edge (-6,0.5);
        \draw[->]  (e2) edge (-3.5,0.5);
        \draw[->]  (e3) edge (-0.5,0.5);
        \draw[->]  (e4) edge (1,0.5);
        \draw[->]  (e5) edge (2.5,0.5);
        \draw[->]  (e6) edge (4,0.5);
        \draw[->]  (e7) edge (5.5,0.5);
        
        \draw [thin, rounded corners, fill=green!25] (-6.65,3.2) rectangle (-5.35,2.8);
              \node [matrix] (o1) at (-6,3) {
              \draw [fill=green] circle (1mm); & \draw [fill=green] circle (1mm); & \draw [fill=green] circle (1mm); & \draw [fill=green] circle (1mm); & \draw [fill=green] circle (1mm); \\
              };
        \draw [thin, rounded corners, fill=green!25] (-4.15,3.2) rectangle (-2.85,2.8);
              \node [matrix] (o2) at (-3.5,3) {
              \draw [fill=green] circle (1mm); & \draw [fill=green] circle (1mm); & \draw [fill=green] circle (1mm); & \draw [fill=green] circle (1mm); & \draw [fill=green] circle (1mm); \\
              };
        \draw [thin, rounded corners, fill=green!25] (-1.15,3.2) rectangle (0.15,2.8);
              \node [matrix] (o3) at (-0.5,3) {
              \draw [fill=green] circle (1mm); & \draw [fill=green] circle (1mm); & \draw [fill=green] circle (1mm); & \draw [fill=green] circle (1mm); & \draw [fill=green] circle (1mm); \\
              };
        \draw [thin, rounded corners, fill=green!25] (0.35,3.2) rectangle (1.65,2.8);
              \node [matrix] (o4) at (1,3) {
              \draw [fill=green] circle (1mm); & \draw [fill=green] circle (1mm); & \draw [fill=green] circle (1mm); & \draw [fill=green] circle (1mm); & \draw [fill=green] circle (1mm); \\
              };
        \draw [thin, rounded corners, fill=green!25] (1.85,3.2) rectangle (3.15,2.8);
              \node [matrix] (o5) at (2.5,3) {
              \draw [fill=green] circle (1mm); & \draw [fill=green] circle (1mm); & \draw [fill=green] circle (1mm); & \draw [fill=green] circle (1mm); & \draw [fill=green] circle (1mm); \\
              };
        \draw [thin, rounded corners, fill=green!25] (3.35,3.2) rectangle (4.65,2.8);
              \node [matrix] (o6) at (4,3) {
              \draw [fill=green] circle (1mm); & \draw [fill=green] circle (1mm); & \draw [fill=green] circle (1mm); & \draw [fill=green] circle (1mm); & \draw [fill=green] circle (1mm); \\
              };
        \draw [thin, rounded corners, fill=green!25] (4.85,3.2) rectangle (6.15,2.8);
              \node [matrix] (o7) at (5.5,3) {
              \draw [fill=green] circle (1mm); & \draw [fill=green] circle (1mm); & \draw [fill=green] circle (1mm); & \draw [fill=green] circle (1mm); & \draw [fill=green] circle (1mm); \\
              };
        \draw[->] (-6,2.5) -> (o1);
        \draw[->] (-3.5,2.5) -> (o2);
        \draw[->] (-0.5,2.5) -> (o3);
        \draw[->] (1,2.5) -> (o4);
        \draw[->] (2.5,2.5) -> (o5);
        \draw[->] (4,2.5) -> (o6);
        \draw[->] (5.5,2.5) -> (o7);
              
        \node at (-8.5,4) {$\pi^g$};
        \node [red_circle] (g3) at (-0.5,4) {\small $\times$};
        \node [green_circle] (g5) at (2.5,4) {\small \checkmark};
        \node [red_circle] (g6) at (4,4) {\small $\times$};
        \draw[->]  (o5) edge (g5.south);
        
        \node at (-8.5,5) {\textsc{\gen{}}};
        \draw [rounded corners, fill=yellow!25]  (1.5,5.25) node (v4) {} rectangle (3.5,4.75);
        \node (gn3) at (1.68,5.4) {\tiny {\texttt{of}}};
        \node (gn5) at (2.5,4.85) {};
        \draw [fill=orange] (1.6,5.16) rectangle (1.75,4.80);
        \draw [fill=orange!40] (1.8,4.96) rectangle (1.95,4.80);
        \draw [fill=orange!40] (2.0,4.80) rectangle (2.15,4.80);
        \draw [fill=orange!40] (2.2,4.90) rectangle (2.35,4.80);
        \draw [fill=orange!40] (2.4,5.05) rectangle (2.55,4.80);
        \draw [fill=orange!40] (2.6,5.00) rectangle (2.75,4.80);
        \draw [fill=orange!40] (2.8,4.85) rectangle (2.95,4.80);
        \draw [fill=orange!40] (3.0,4.83) rectangle (3.15,4.80);
        \draw [fill=orange!40] (3.2,4.86) rectangle (3.35,4.80);
        \draw[->] (g5) edge (gn5.south);
        
        \node at (-8.5,6) {\textsc{\detok{}}};
        \node (gn3) at (-0.5,6) {\tiny {\texttt{i~~s~~/w}}};
        \draw[->,>=to] (-0.95,5.8) to [bend right] (-0.6,5.8);
        \draw[->,>=to] (-0.5,5.8) to [bend right] (-0.15,5.8);
        \node (gn6) at (4,6) {\tiny {\texttt{e~~e~~l~~s~~/w}}};
        \draw[->,>=to] (3.15,5.8) to [bend right] (3.5,5.8);
        \draw[->,>=to] (3.55,5.8) to [bend right] (3.9,5.8);
        \draw[->,>=to] (4,5.8) to [bend right] (4.35,5.8);
        \draw[->,>=to] (4.4,5.8) to [bend right] (4.8,5.8);
        \draw[bend left, ->] (o3) edge (gn3.south west);
        \begin{pgfonlayer}{background}
              \draw[bend left, ->] (o6) edge (gn6.south west);
        \end{pgfonlayer}
    
    \end{tikzpicture}
    
    \caption{\tokdetok{} integration into a masked language model. $\pi^t$ is set to send all multi-\wpc{} words into \tok{}; $\pi^g$ is set to generate every third word using \detok{}. {\texttt{/w}} is a reserved end-of-word character.}
    \label{fig:algo}
\end{figure*}
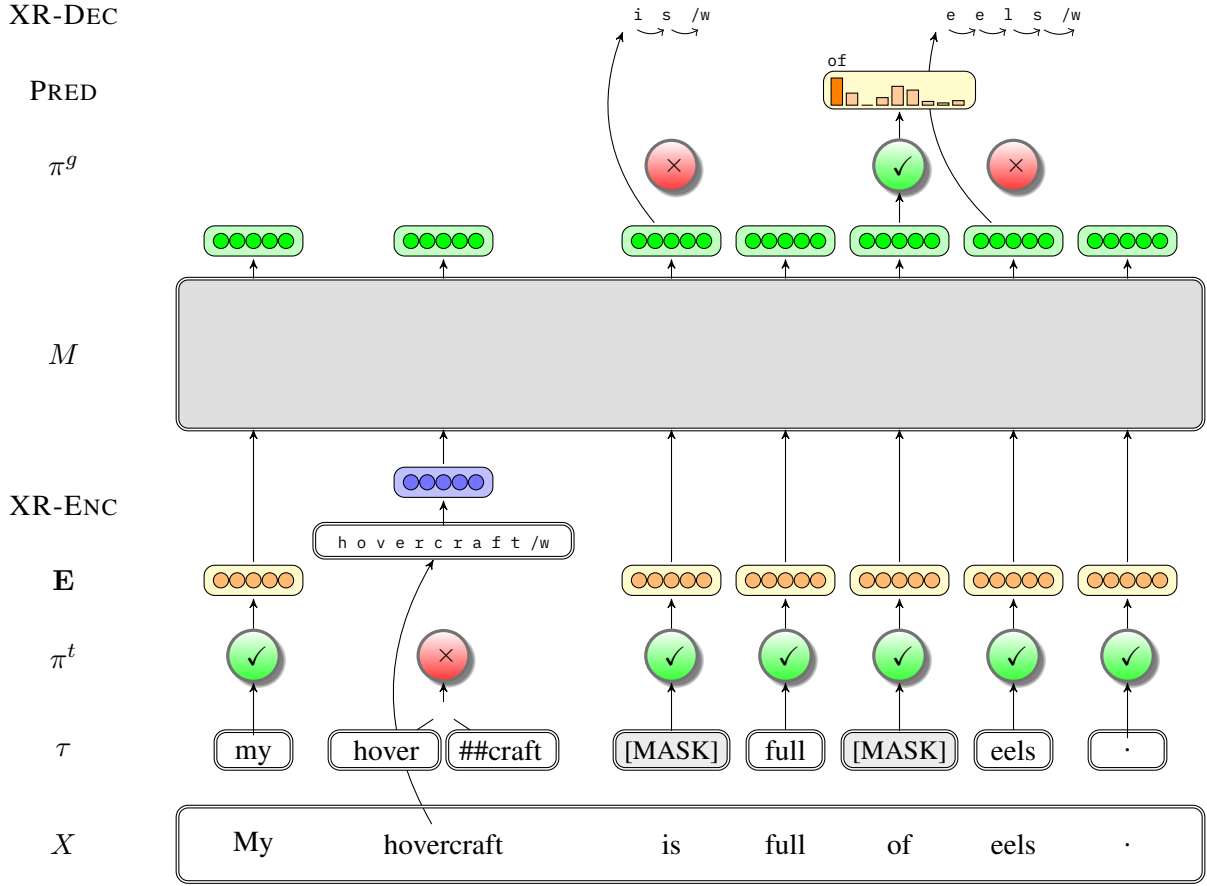

When shifting between domains, topics, and registers, this property produces undesired consequences~\cite{blitzer-etal-2007-biographies}.
One effect of training or fine-tuning an \llm{} over a dataset extracted from a domain other than the one on which its subword vocabulary was created is the formation of long input sequences with different words receiving markedly different treatment.
For example, in the tweet shown in \autoref{tab:example}, taken from a task where an emoji is to be matched to a tweet (see \S\ref{sec:tasks}), the intended emoji is the US flag; 
but the excessive number of \wpc{} tokens, and the breaking down of the most informative word \say{\#4thofjuly} into four unrecognizable segments, including the splitting of \say{July} due to its mid-word location, causes a classifier trained over a fine-tuned BERT model to falsely predict the red heart emoji. 
In sequence labeling tasks like named entity recognition or morphological tagging, the breaking down of sequence-atomic words into tokens presents a nontrivial decision point as to how multi-token words' output representations are to be used for predicting labels~\cite{acs2021subword}.
In languages with non-concatenative morphology such as Hebrew, the contiguous nature of subword tokens has been shown to render the representations of the transformer stack all but useless~\cite{klein-tsarfaty-2020-getting}.

In this paper we present \tokdetok{}, a system which provides \llm{}s the option to represent novel and rare words with a single vector composed over their character sequence, while retaining the knowledge learned for well-recognized word tokens during pre-training.
Our method bridges the gap between the two representation modules via an additional pre-training sequence where the language modeling objective is supplemented with training a character-level encoding component which provides vector inputs to the contextualized model apparatus, as well as a generative component which learns to output target words from their contextualized representations.
We show how these lightweight auxiliary components can be trained so that they regularize each other, resulting in a character-to-vector model that approximates representations well for single- and multi-token words, and in a word-generation model which produces well-formed sequences approximating English from arbitrary points in vector space.
We examine the performance of various \llm{}s trained with a \tokdetok{} module over text from both user-generated and edited sources on both sequence classification and sequence labeling tasks, finding that our system's strength lies mostly in the latter.
We then analyze the effect of individual components and decision points in our system, such as our choice of corpus for the second pre-training phase and our regularization regime.\footnote{We will release our code and model checkpoints.}

\section{\tokdetok}
\label{sec:model}

We focus on character-informed representations for \llm{}s which use the transformer architecture~\cite{vaswani2017attention}, such as BERT~\cite{devlin-etal-2019-bert}, RoBERTa~\cite{liu2019roberta}, 
and GPT~\cite{radford2018improving}.
A Transformer \llm{} is composed around a core module \mmod{} parameterized as multi-head self-attention layers, which accepts a sequence $\{\ve_1, \dots, \ve_l\}$ of embedding vectors corresponding to a list of token indices $\{t_1, \dots, t_l\}$ from a vocabulary $\mathcal{V}$,
and outputs a sequence of contextualized vectors $\{\vh_1, \dots, \vh_l\}$, of which a subset $\{\vh_i\}_{i\in\mathcal{I}}$ correspond to masked tokens in positions $\mathcal{I}\subseteq [l]$ selected by a boolean masking operator $m$.\footnote{In the case of an autoregressive \llm{} like GPT, $\mathcal{I} = [l]$ but each vector $\vh_i$ is only dependent on the tokens preceding its position, $\{t_1, \dots, t_{i-1}\}$, and token-wise prediction is performed for each position assuming no knowledge of future tokens.}
The token index list supplied to \mmod{} is the output of a tokenization function $\tau$ operating on a sequence $X$ followed by lookup in an embedding table $\vE$,
while its output vectors $\{\vh_1, \dots, \vh_l\}$ serve as input to a prediction module \gen{}, which outputs a distribution $D_i$ over $\mathcal{V}$ for each masked position $i$.
All embeddings $\{\ve_i\},\{\vh_i\}$ live in a shared space $\mathds{R}^d$.
Together, the components described so far operate in the following manner:

\begin{equation}
    \{D_i(\mathcal{V})\}_{i\in\{\mathcal{I}\}} = 
    \text{\gen{}}\left( \text{\mmod{}} \left( \vE\left[ \tau(X) \right] \right) \right),
\end{equation}
where square bracketing denotes elementwise table lookup.

In the considered \llm{} architectures, the input $X$, which is atomically made up of a sequence of characters $\{c_1, \dots, c_n\}\in\Sigma^{n}$, is broken down by $\tau$ to provide the tokenization of length $l\leq n$,
and the prediction/generation operator \gen{} accepts the contextualized outputs $\{\vh_1, \dots, \vh_l\}$ and implements prediction by means of a softmax distribution which is based on scores obtained via dot-product against an output embedding table, which is usually the same as $\vE$.

The \tokdetok{} model makes no adjustments to \mmod{} itself, and only offers conditional replacements for $\tau$ and \gen{}.
An \textbf{encoding policy} $\pi^t$ selects a subset of tokens from the original sequence $\mathcal{J}^t\subseteq [l]$ to be represented by \tok{} instead of $\vE \circ \tau$.
\tok{} has access to the part of the character sequence $X$ which underly the tokens selected by $\pi^t$, and produces input embeddings directly from the character level.
These alternate embeddings must agree in dimension with those of each $\ve$, but a single embedding may be used to replace multiple base tokens (usually when all tokens corresponding to a single out-of-$\vE$ word are replaced), resulting in a shorter input sequence for \mmod{}.
A separate \textbf{generation policy} $\pi^g$ selects a subset of the original sequence $\mathcal{J}^g\subseteq [l]$ to be generated from the output vectors $\vh$ by \detok{} instead of \gen{}.
A high-level schematic depicting this framework is presented in \autoref{fig:algo}.

\begin{table*}
    \centering
    \tiny
    \begin{tabular}{ll}
        \toprule
        {\small Original} & \texttt{He was emphatically~~~~~a modern gentleman,~~of scrupulous~~~~~~~courtesy,~~sportive~~~gaiety,} \\ 
        {\small Word pieces} & \texttt{He was em \#pha \#tically a modern gentleman , of s \#c \#rup \#ulous courtesy , sport \#ive g \#ai \#ety ,} \\
        \midrule
        {\small Random-20\%} & \texttt{He was [TOK]~~~~~~~~~~~~a [TOK]~~gentleman , of s \#c \#rup \#ulous~[TOK]~~~~, sport \#ive g \#ai \#ety ,} \\
        {\small All-multi} & \texttt{He was [TOK]~~~~~~~~~~~~a modern gentleman , of [TOK]~~~~~~~~~~~~courtesy , [TOK]~~~~~~[TOK]~~~~~~,} \\
        {\small \textsc{Suffixes}} & \texttt{He was [TOK]~~~~~~~~~~~~a modern gentleman , of [TOK]~~~~~~~~~~~~courtesy , sport \#ive [TOK]~~~~~~,} \\
        \bottomrule
    \end{tabular}
    \caption{Example of token input selection by different policies,
    where \texttt{[TOK]} signifies a word to be replaced by its character-based representation from \tok{}.
    Sentence fragment taken from the MS-MARCO QA dataset (see \S\ref{sec:tasks}) and tokenized using BERT-cased (\#\# replaced with \# to fit paper width).}
    \label{tab:policies}
\end{table*}

The specific policies in a given application may be defined based on the model's use case.
For example, in text classification no generation is required, and so $\pi^g$ will return $\emptyset$ for all sequences; $\pi^t$ can be tuned for a task based on known features of the base model (BERT/GPT etc.) and of the domain text, some examples including:
tokens corresponding to all words that are not in a pre-determined vocabulary;
all words in the sequence;
all words assigned more than one token by the base tokenizer $\tau$;
a random sample of words in the sequence;
or all words including characters that are not lowercase English characters.
One particular policy we hypothesize could be useful is one that affords the tokenizer slack in detecting a single simple derivational or inflectional suffix: all words which are single-token or whose second-and-final token is in the list \textsc{Suffixes} are left for $\vE \circ \tau$; the rest are represented using \tok.\footnote{\textsc{Suffixes}, compiled by manually examining a list of most common second-and-final tokens in a large corpus under GPT-2's tokenization, = $\{$\emph{s}, \emph{ed}, \emph{es}, \emph{ing}, \emph{ly}, \emph{al}, \emph{ally}, \emph{'m}, \emph{'re}, \emph{'ve}, \emph{y}, \emph{ive}, \emph{er}, \emph{'t}, \emph{'ll}, \emph{an}, \emph{ers}$\}$.
Similar policies can be hand-crafted for different languages, based on learner-level knowledge of the language and minimal preliminary analysis of sample tokenizations. Our results show they are not strictly necessary.}
Different policies may be applied in training settings as well, for example in order to \say{familiarize} the heavily-parametrized \mmod{} with the inputs from \tok{}.
Three example policies are illustrated in~\autoref{tab:policies}.

\subsection{Second Pre-training}
\label{ssec:pretr}

A typical \llm{} is initiated through a computationally-intensive pre-training step, iterating over a large corpus in batches of sequences and backpropagating a cross-entropy loss calculated over the prediction layer's output into all of its components, through \gen{} to \mmod{} to $\vE$.
In order to train \tokdetok{}, we introduce a second pre-training step we term \ppt{}, where the \llm{} continues to update its parameters for a (possibly different) corpus, but is supplemented with the \tokdetok{} elements in order to \say{acclimatize} the \mmod{} components to outputs from \tok{}.\footnote{Within the taxonomy of phases between pre-training and task training~\cite{ruder2021lmfinetuning}, this is closest to \textbf{Adaptive Fine-Tuning}. However, due to our added modules we opt to assign it a new name.}
In addition, \tok{} is also trained through a lower-level objective requiring it to approximate the outputs of $\vE \circ \tau$ which it is replacing, and \detok{} is trained to sequentially produce the correct character sequence from \mmod{}'s outputs.
Together, a batch of text sequences in a \ppt{} step produces the following loss elements which are backpropagated into the unified model:
\begin{itemize}
    \item A \textbf{language modeling loss} from the softmax operation over masked tokens, updating \mmod{}'s and $\vE$'s parameters, as well as \tok{}'s for tokens selected by a usage policy $\pi^t_{(u)}$ to be used in \mmod{}'s input;
    \item An \textbf{embedding loss} for the \tok{} component, computed against $\vE \circ \tau$'s token embeddings, over a set selected by a policy $\pi^t_{(l)}$ which may or may not equal $\pi^t_{(u)}$.
    This loss can be computed, e.g., as the euclidean distance between the output and the target.
    When the target corresponds to multiple token counts, an aggregation pooling function $\agg:\mathds{R}^{d \times \mathds{N}^{+}}\rightarrow\mathds{R}^d$ needs to be defined over the embeddings, for example taking their dimension-wise mean, taking the leftmost token's embedding, or taking the dimension-wise $\max$;
    \item A \textbf{generation loss} for words generated by \detok{} from \mmod{}'s output vectors, according to a $\pi^g$ policy.
    This loss is the character-level cross-entropy for autoregressive sequence generation.
    Note that in order to only generate full words, $\pi^g$ must align with $\pi^t_{(u)}$ so that no multi-token words left as input to \mmod{} are also selected for generation.
\end{itemize}

As a form of regularization within the \tokdetok{} components, we introduce additional training batches we call \textbf{cycle dependency loops}.
In such a batch, \tok{} and \detok{} act in succession, starting from one of the spaces they operate in, with the goal of arriving at the same point after cycling through both components.
An \textbf{\tdloop} loop thus starts at a character sequence $\vc\in{\Sigma^*}$, runs it through \tok{} to obtain a vector $\ve=\tok(\vc)$, and runs \detok{} in an attempt to return to the original sequence $\hat{\vc}=\detok(\tok(\vc))\approx \vc$.
Analogously, a \textbf{\dtloop} loop starts at a vector $\tilde{\ve}\in\mathds{R}^{d}$ and targets $\hat{\ve}=\tok(\detok(\tilde{\ve}))\approx \tilde{\ve}$.
In this loop, loss is only backpropagated as far as \tok{}, since backpropagating through a generative model's decision component introduces discrete steps which must be smoothed or approximated (see discussion in \newcite{peng-etal-2018-backpropagating}).

\section{Tasks}
\label{sec:tasks}

To evaluate the advantages of character-sequence awareness in large-scale transformers, we chose a diverse set of datasets which reflect unedited user-generated language in English, as well as its interaction with edited text.
We report results on a sequence classification task (emoji prediction), a sequence tagging task (named entity recognition, or NER) in both an in-domain (Twitter NER) and cross-domain (emerging entities NER) setting, a sequence ranking task based on information retrieval in a hybrid edited-unedited textual setting (MARCO-QA ranking), and a task where a single word's class is predicted within a sequence (NYTWIT).

\paragraph{Emoji prediction.}
The English portion of the Multilingual Emoji Prediction dataset \cite{barbieri-etal-2018-semeval,ma-etal-2020-multi} is composed of tweets containing one of the twenty most common emoji symbols.
For the task, the emoji are stripped from the tweet text and the system is asked to predict which one appeared in the original tweet, allowing it to be construed as a self-annotated, fine-grained sentiment analysis task.
All tweets are identified as geographically originating in the US.
We note that there is a significant qualitative difference between the training (+ development) set, and the test set of this corpus, hinted at by one of the participant teams in the original task~\cite{chen-etal-2018-peperomia} but not explored.\footnote{Another team identified a seasonal shift affecting the distribution of the Christmas tree 
emoji~\cite{coster-etal-2018-hatching}.}
The partitions were extracted based on a temporal split, with all test set tweets post-dating all training set tweets by at least three months. 
More crucially, the test set dates (May 2017 -- January 2018) overlap with Twitter's increase of the tweet character limit from 140 to 280 characters, phased in mostly during November 2017.
As a result, test set tweets have on average $\sim$10\% more words, increasing the amount of information within and diverging their textual feature distribution from that of the training set by more than is customary in NLP tasks.
The label distribution remains more or less the same.

\paragraph{Twitter NER.}
The Twitter NER dataset \cite{strauss-etal-2016-results} is a prime example of sequence tagging in noisy user-generated data settings.
It is composed of randomly sampled English tweets from 2016, annotated for ten different entity types, including music artists and sports teams, thus representing topics prevalent in social media.

\paragraph{Emerging entities recognition.}
The emerging entities dataset \cite{derczynski-etal-2017-results} shares most of its training set with the same Twitter source as the Twitter NER dataset, but as a domain-adaptation setup it includes a more ambitious evaluation fold: the development set is extracted from YouTube comments, and the test set from StackExchange and Reddit.
The taskmasters claim that the dataset contains mostly rare and hitherto-unseen entities, but do not provide exact statistics.

\paragraph{Community question answering.}
The MS-MARCO question answering dataset~\cite[\marco;][]{nguyen2016ms} was collected by mining a commercial search engine log for user queries and asking humans to answer them, supplying the answer writers with a set of passages retrieved automatically from edited text by the search engine which the answer writers then marked as \say{selected} if they helped them formulate the answer.
We recast this dataset into a selection / ranking problem, not pursued in the original challenge: given the query and the set of possibly helpful passages, which is the passage the answer writer selected?
To this end, we filtered out queries with more or fewer than one selected passage, and evaluated each system based on the mean reciprocal ranks (MRR) of the true selected passages in the rankings it produced.
Due to its size, we uniformly sampled 10\% of the queries and associated passage collections from each partition in this dataset, and ran all analyses and experiments on this new sampled dataset.

\paragraph{Novel word classification.}
The NYTWIT dataset~\cite{pinter-etal-2020-nytwit} includes passages in the New York Times where a word appears which has not appeared in the publication before.
The system is tasked to classify the novel word into one of eighteen types of novelty sources, such as \say{inflection of known word} or \say{lexical blend}.
The dataset is not partitioned into train/dev/test, and so we use the 10-fold partition from \newcite{pinter-etal-2020-nytwit} and report accuracy results aggregated over all instances, each from a model trained on the other nine folds.

\begin{table*}
    \centering
    \small
    \begin{tabular}{lcccccc} \toprule
        Dataset & Instances & Tokens & Types & TTR & Multitoks & Token mass \\
         & & \multicolumn{2}{c}{(Space-delimited)} & & & increase \\
        \midrule
        CoNLL-2003 NER & 14,986 & 204,563 & 23,624 & .115 & 16.08\% & 29.50\% \\
        Wiki1 & --- & 204,564 & 28,092 & .137 & 8.64\% & 12.85\% \\
        \midrule
        NYTWIT & 1,903 & 94,403 & 26,028 & .276 & 24.04\% & 38.04\% \\
        Wiki2 & --- & 94,403 & 16,903 & .179 & 9.11\% & 13.75\% \\
        \midrule
        Twitter NER & 2,394 & 46,469 & 10,586 & .228 & 17.23\% & 43.21\% \\
        Wiki3 & --- & 46,509 & 10,855 & .233 & 9.02\% & 13.77\% \\
        \midrule
        Emerging NER & 3,394 & 62,730 & 14,878 & .237 & 19.15\% & 54.25\% \\
        Wiki4 & --- & 62,757 & 13,392 & .213 & 9.05\% & 13.74\% \\
        \midrule
        Emoji Prediction & 427,458 & 4,973,813 & 504,644 & .101 & 32.96\% & 64.90\% \\ 
        Wiki5 & --- & 4,973,813 & 194,240 & .039 & 9.07\% & 13.58\% \\ 
        \midrule
        \marco{} (10\% sample) & 80,704 & 46,956,674 & 1,355,049 & .029 & 20.58\% & 31.91\% \\ 
        Wiki6 & --- & 46,956,703 & 818,712 & .017 & 9.48\% & 14.37\% \\
        \bottomrule
    \end{tabular}
    \caption{Surface statistics for the datasets used for evaluation (training sets), against Wikipedia text of comparable token count sizes. Tokenization performed with GPT-2.}
    \label{tab:stats}
\end{table*}

\subsection{Analysis}
\label{ssec:stats}

We begin with an analysis of the datasets and their subword properties in order to gauge the direct effects of tokenization on the dataset's genre and domain.
In \autoref{tab:stats}, we present surface-level statistics reflecting the challenges posed by the datasets' sources, comparing each task's training set with comparably-sized sets of sentences sampled uniformly from English Wikipedia.
We use the GPT-2 tokenizer, which boasts the largest subword vocabulary of all models considered in this work, to obtain a lower bound on the added token mass expected by our models on these datasets, and report the following measures: the number of unique word types and type/word ratio, the percent of types which are subword-tokenized by GPT-2 (\textit{Multitoks}), and the overall increase in number of tokens over the corpus compared to a single-token-per-word representation (i.e., strictest application of \tokdetok{} considered in this work).

We find a striking disparity between the well-edited Wikipedia corpora, themselves far from being completely in-vocabulary, and the datasets at hand.
Wikipedia itself proves to be scale-invariant on the metrics that are not TTR, maintaining a token-OOV rate of roughly 9\% and a tokenization overhead of $\sim$13\%.\footnote{This number is very close to the one found by \newcite{acs2021subword} using multilingual transformer models; other languages' corpora boast overheads ranging from 28\% (French) to 95\% (Japanese).}
A control dataset in the form of CoNLL-2003's English NER portion~\cite{tjong-kim-sang-de-meulder-2003-introduction}, extracted from newswire text, exhibits a marked increase in complex words, possibly mostly named entities, or time-sensitive terms compared to the tokenizer's training set, alongside a decrease in type count which reflects its narrow source domain.
NYTWIT, despite also being extracted from newswire text, is sourced from news items that contain novel forms and so are often written in a high register or involve niche domains; as a result, it contains a larger overall token mass, split rather evenly across individual words (i.e., unknown words have fairly \say{regular} structure).
The Twitter NER datasets, exhibiting tweet language, do not exceed CoNLL's multitoken type proportion by much, but its OOVs tend to be completely unexpected forms, leading to a much higher raw post-tokenization count.
In the emoji dataset, which has not been pre-processed according to NER standards and instead was directly scraped off Twitter, almost a third of all unique forms are multi-token, and their presence enlarges the total token count by nearly two thirds.
\marco{} data, most of which is text from highly-ranked web pages, but also including user-generated queries, assumes a middle position between these two extremes.

\section{Experiments}
\label{sec:exp}

We evaluate the effect of \tokdetok{} when included in various \llm{}s. 
We choose \textsc{BERT-base-cased}~\cite{devlin-etal-2019-bert}, \textsc{GPT2-small}~\cite{radford2019language}, and \textsc{RoBERTa-base}~\cite{liu2019roberta} as the base \llm{}s to be manipulated.\footnote{Models loaded from the Huggingface repository~\cite{wolf-etal-2020-transformers}.}
The former contains roughly 108M parameters, and the latter two roughly 125M, a difference accounted for by their larger subword vocabulary (50k vs. 29k) and the resulting larger embedding table.
All models are case sensitive, but they differ in their strategy for preserving the original space-delimited word sequence: BERT's tokenizer marks word-non-initial tokens with a \say{\#\#} string, while GPT-2's tokenizer marks spaces with a special underline character and appends them to the following word.
The difference manifests itself in sequence-initial words, whose initial token in the GPT-2 representation shares the form of sequence-medial word-medial tokens, rather than that of sequence-medial word-initial tokens; and in symbols pre-tokenized without a preceding space, such as punctuation and apostrophes.
RoBERTa's tokenizer adopts the GPT-2 marking strategy but avoids the first pitfall by internally prepending all input sequences with a space character.
We perform the \ppt{} phase for each model on a collection of English tweets from 2016 obtained from the Firehose and preprocessed to replace all `@'-mentions with \texttt{@user}.
We later ablate this domain change effect by training models with \ppt{} text from the English Wikipedia March 2019 dump (see \S\ref{ssec:abl}).
We sample both resources to create pre-training corpora of roughly 725MB (unzipped), several orders of magnitude smaller than what contemporary models use for the first pre-training phase, and train for a single epoch.

In preliminary experiments on several character-level \tok{} architectures, we found that a convolutional net outperforms bidirectional LSTMs and small transformer stacks.
We pass the input characters through three separate convolution layers of width 2, 3, and 4 (characters), then pass the outputs through max-pooling layers and a \texttt{ReLU} activation, and finally project the concatenation of the results onto the base models' embedding dimension.
This \tok{} component contains $\sim$1M parameters, negligible compared to the transformers' parameter count.\footnote{The Wikipedia-trained models contain $\sim$2.8M parameters, the difference owing to language-ID filtering performed on the Twitter data, leading to a much smaller character set.} 
We implement \detok{} as a 2-layer unidirectional LSTM whose hidden layer is initialized by projecting the context vector $\vh_i$ output from \mmod{} into the hidden dimension.
Characters are generated by projecting the LSTM's output through two linear layers with a \texttt{tanh} activation.

During \ppt{}, we insert a cycle dependency batch every 5,000 LM steps.
For the replacement policies we choose $\pi^t_{(l)}$ to sample uniformly random sets of tokens representing 15\% of space-delimited words to pass as a target loss
for \tok{}, and $\pi^t_{(u)}$ to replace the embedding input to \mmod{} with \tok{}'s output for all multi-token words.
$\pi^g$ selects all tokens to pass as target losses for \detok{}, calculated as a sum of the cross-entropy loss for each character in the target sequence.
Cycle batches consist of sampling $k$ words out of the $K$ most frequent words from the training corpus, with replacement, and $k$ vectors from a Gaussian distribution blown to concentrate around the surface of the unit sphere in hidden-dimension space:
\[ \tilde{\ve} \sim \frac{1}{\sqrt{d}}\cdot\mathcal{N}(0,\vI^{(d)}). \]
\tdloop{} loops are optimized for a
character-level cross-entropy
loss, whereas \dtloop{} loops target a
euclidean distance loss.
When a \tok{} embedding correspond to multiple $\tau$ tokens, the learning target is created by max-pooling their embeddings.\footnote{This $\agg$ function outperformed average-pooling and first-token selection in preliminary experiments.}
We set $K=$25,000 and $k=$1,000.

\subsection{Intrinsic Assessment}

\begin{figure}
    \centering
    \includegraphics[width=8.25cm]{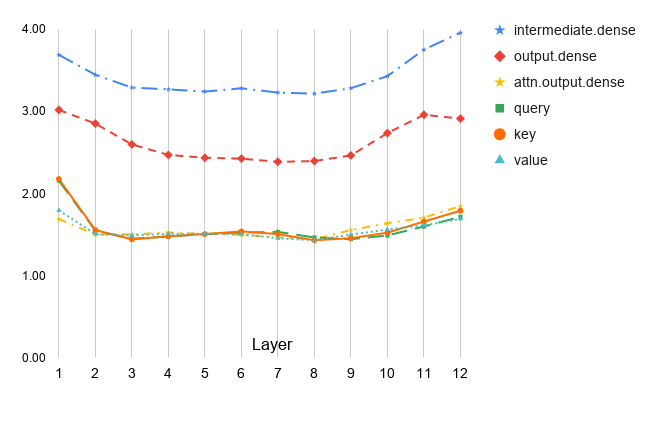}
    \caption{Euclidean distance between RoBERTa weight parameter values before and after a \ppt{} training phase on the Twitter corpus.}
    \label{fig:updates}
\end{figure}

Across all base models and both \ppt{} corpora selected for our experiments, we observed a steady decrease in the \llm{} models' built-in loss metrics (masked prediction / autoregressive prediction) until stabilizing at roughly half the initial value before the end of the \ppt{} epoch.
This indicated that the transformer layers are able to process inputs from both the embedding table and \tok{} and reconcile them.
\autoref{fig:updates} depicts the parameter updates in RoBERTa by parameter type, across layers, comparing parameter values before and after the \ppt{} phase on Twitter data.
It shows that the change along the model layers is fairly stable, with mildly more extensive updates in the bottom and top layers.
The former is to be expected given the introduction of inputs from \tok{}; the latter can also be influenced by encountering Twitter data, which is substantially different than what RoBERTa is \say{used to}.

\begin{table*}
    \centering
    \footnotesize
    \begin{tabular}{lHcccccc}
        \toprule
        Model & Uncased BERT & BERT & \multicolumn{2}{c}{GPT-2} & \multicolumn{2}{c}{RoBERTa} \\ 
        Corpus & Wikipedia & Wikipedia & Wikipedia & Twitter & Twitter & Twitter \\
        Steps & 76,000 & 9,000 & 13,500 & 21,000 & 7,500 & 57,500 \\
        Sequences ($10^3$) & 3,648 & 5,184 & 7,776 & 10,080 & 2,160 & 16,560 \\
         \midrule
        & ppercented & proming & crordman & d & orereren & everyone \\
        & or & dy & sssion & . & ant & kerned \\
        & ter & deded & gental & the & re & levernger \\
        & peprepored & terse & 2 & @ & cerent & and \\
        & essed & h & ther & ==666!!!!!!!!!!!!!!! & ennte & ed \\
         \bottomrule
    \end{tabular}
    \caption{Example generated words from random locations near the surface of the unit sphere in $\mathds{R}^{768}$.}
    \label{tab:gen}
\end{table*}



Another artifact of \ppt{} is the outputs of the \detok{} module.
While monitoring the training procedure, we periodically sample words from random locations centered around the surface of the unit sphere of the embedding space, to see what \say{priors} the generative net is learning from the vectors encountered during training.
\autoref{tab:gen} presents some of these samples for different models and different corpora at different points in the training phase.
Masked models appear to be learning well-formed fractions of English or pseudo-English words at early stages of the training phase from both Wikipedia and Twitter data.
As training continues, fewer sequences containing repetitions are observed, fewer generations occur across  samples, and more in-vocabulary words appear, suggesting convergence of the vector space towards representing well-formed and diverse English vocabulary.
The autoregressive GPT-2, on the other hand, struggles to produce meaningful sequences beyond short words and punctuation symbols when trained on the informal Twitter input, suggesting a difficulty in learning a mapping of language from vector space without availability of a two-sided context.

\subsection{Downstream Evaluation}
\label{ssec:finetune}

During task fine-tuning and inference,
since we do not evaluate on generative tasks, we do not use \detok{}.
We perform minimal hyperparameter search for each base model + task combination, and fine-tune model parameters during downstream training in all tasks but NYTWIT.
For these tasks, we also experimented with setups where \mmod{} and \tok{} are used as feature extractors only, and where \tok{} training is supplemented by an additional embedding loss (as described in \S\ref{ssec:pretr} for the \ppt{} phase) computed against embeddings of the task input.
Neither setup provided improvement on any task during our tuning experiments (see \S\ref{ssec:abl}).

Downstream models are implemented as follows: for sequence classification and ranking, a two-layer perceptron with \texttt{ReLU} activation is trained to make the prediction from the top-layer representation of the initial \texttt{[CLS]} token (in BERT and RoBERTa models) or of the final token in the sequence (in GPT-2).
For NER, an LSTM is run over the sequence of each word's top-layer representation, followed by a single linear layer which makes the prediction.
For NYTWIT, \mmod{}'s contextualized vector for the target word is used as input for a single logistic layer.
In cases of multi-token words, the prediction from the first token is selected.
We set \tok{}'s character embedding dimension to 200 and the convolutional layers to 256 channels.
Following \newcite{sun2019fine}, we set the maximum learning rate to $10^{-3}$ for the task models and $2\times 10^{-5}$ for fine-tuning, and perform warm-up for 10\% of the total expected training steps before linearly decaying the rates to zero.
All parameters are optimized using Adam~\cite{adam} with default settings.
We run all NER models for twenty epochs and sequence-level task models for three, evaluating on the validation set after each epoch using the metrics reported below, and stopping early if performance has not improved for four epochs.
In order to avoid unfairly favorable conditions for \tokdetok{} models, task hyperparameters are all tuned on the base models, with \tokdetok{} models using only values on which their base equivalents have also been evaluated.

We evaluate the effectiveness of \tokdetok{}'s concepts and components by comparing the following setups:\footnote{A setup where all words are represented by \tok{} obtained noncompetitive results on all tasks.}
\begin{itemize}
    \item \textsc{\textbf{None}} uses only the base model;
    \item \textsc{\textbf{None+\ppt{}}} uses a version of the base model that was further pre-trained on the same Twitter corpus on which \tokdetok{} is trained (adaptive fine-tuning~\cite{ruder2021lmfinetuning}), in order to control for the increase in total unlabeled text seen by the model;
    \item \textsc{\textbf{Scaffolding}} is a model trained with \tokdetok{} in a \ppt{} phase, but only using base model embeddings during task fine-tuning;
    \item \textsc{\textbf{Stochastic}} samples 10\% of the words in the downstream datasets, calling \tok{} on their character sequences while using the base model's embedding(s) for the remaining 90\%;
    \item \textsc{\textbf{All no-suff}} calls \tok{} on all multi-token words which are not of the form \texttt{[token suff]}, where \texttt{suff} is a member of the \textsc{Suffixes} set described in \S\ref{sec:model}, and uses the base model's embedding on the rest.
\end{itemize}

\begin{table*}
    \centering
    \small
    \begin{tabular}{llccccccccc}
        \toprule
        Base & \tokdetok{} & \multicolumn{2}{c}{Emoji} & \multicolumn{2}{c}{Twitter NER} & \multicolumn{2}{c}{Emerging NER} & QA & NYT- & Avg. \\
        & & Dev & Test & Dev & Test & Dev & Test &  & WIT & Test \\
        \midrule
        BERT & None & 24.30 & 37.29 & 29.06 & 28.66 & 38.14 & 29.01 & 46.78 & 45.21 & 37.39 \\
        & None+\ppt{} & \underline{\textbf{29.25}} & \textbf{39.40} & 32.05 & 29.50 & 39.79 & \textbf{29.25} & \underline{\textbf{50.02}} & 44.87 & 38.61 \\
        & Scaffolding & 28.17 & 38.60 & 32.45 & \textbf{31.38} & \textbf{41.10} & \textbf{29.25} & 49.50 & \textbf{53.62} & \textbf{40.47} \\
        & Stochastic & 27.54 & 37.41 & 30.77 & 27.51 & 37.99 & 27.98 & 48.12 & 49.80 & 38.16 \\
        & All no-suff & 27.12 & 38.63 & \textbf{34.09} & 29.59 & 39.47 & 28.43 & 48.64 & 34.76 & 36.01 \\
        \midrule
        GPT2 & None & 24.89 & 38.72 & 30.25 & 26.70 & 40.06 & 28.40 & 44.90 & 47.25 & 37.19 \\
        & None+\ppt{} & 25.47 & \textbf{40.95} & 32.00 & 28.51 & 40.68 & 29.90 & \textbf{47.18} & 47.44 & 38.80 \\
        & Scaffolding & 25.29 & 40.70 & 30.46 & 28.97 & 41.16 & 28.23 & 46.99 & \textbf{50.99} & \textbf{39.18} \\
        & Stochastic & 25.23 & 38.55 & 32.20 & 28.10 & 39.58 & 28.35 & 46.11 & 46.37 & 37.50 \\
        & All no-suff & \textbf{26.30} & 34.27 & \textbf{36.27} & \textbf{32.48} & \textbf{49.20} & \textbf{34.87} & 44.98 & 33.21 & 35.96 \\
        \midrule
        RoBERTa & None & 25.07 & 39.50 & 48.57 & 44.86 & 56.43 & \underline{\textbf{46.22}} & 45.12 & 48.31 & 44.80 \\
        & None+\ppt{} & \textbf{27.04} & \underline{\textbf{42.82}} & 47.39 & 43.84 & 57.13 & 45.12 & \textbf{48.98} & 48.43 & 45.84 \\
        & Scaffolding & 25.87 & 41.12 & \underline{\textbf{49.71}} & \underline{\textbf{45.51}} & 56.28 & 44.60 & 48.26 & \underline{\textbf{53.98}} & \underline{\textbf{46.69}} \\
        & Stochastic & 26.38 & 40.09 & 49.57 & 44.82 & \underline{\textbf{58.21}} & 45.26 & 48.54 & 51.80 & 46.10 \\
        & All no-suff & 26.82 & 33.07 & 46.55 & 42.72 & 55.07 & 43.91 & 47.45 & 35.65 & 40.56 \\
        \midrule
        \multicolumn{2}{l}{SOTA (reported)} & & 47.46 & & 52.4\phantom{0} & & 49.6\phantom{0} & & 48.4\phantom{0} & \\
        \bottomrule
    \end{tabular}
    \caption{Results on all models (Emoji, NER, NYTWIT: Micro-F1 $\times$ 100; QA: MRR $\times$ 100), all results except for \textsc{None} are averaged over three models initialized on different random seeds.
    Best result for each base model in bold, best across all underlined.
    }
    \label{tab:main_results}
\end{table*}

We present the results of the downstream prediction tasks in \autoref{tab:main_results}.
All transformer models except for \textsc{None} were 2nd-phase pre-trained three separate times using different random seeds, and the mean results are reported.

The first observation we make is the dominance of RoBERTa, a thoroughly optimized masked language model, over the other models on the NER datasets in all its variants.
This suggests RoBERTa has captured fine-grained information about individual words that it was able to retain in its representations for them; the struggle of the \tokdetok{} model to provide improvement over the \textsc{None} versions of the model strengthens this hypothesis.
GPT-2, despite having access to only left-side context of each word, still outperforms the basic BERT model on most setups in the NER tasks, perhaps due to its larger subword vocabulary size.
At the same time, its gains from the \tok{} representations are much more considerable, suggesting that its left-context-only inference may in fact be detrimental to the its \mmod{}'s performance as a whole.
In general, models perform better on Emerging NER than on Twitter NER, which we attribute to several possible causes or their combination: first, the source shift in the Emerging NER test set from Twitter to Reddit is meant to encumber the models, but given the extensive pre-training they undergo they may actually benefit from the fact that test sequences are longer on average than those in the Twitter NER dataset;
second, more prosaically, the Emerging NER dataset contains fewer entity types (6 vs. 10), making the task itself somewhat easier.

The other word-level task, NYTWIT, demonstrates substantial gains made by the \tokdetok{} training regime: the \textsc{Scaffolding} setup preforms best in all three base models, and in both masked langugage models the \textsc{Stochastic} setup outperforms both \textsc{None} variants.
Together with the NER results, this suggests that \tokdetok{} succeeds in providing transformer models with \textbf{word-level} representations that allow coarse-grained classifications (such as named entity type or novel word origin), better than default subword segmentations, for both edited and user-generated text.

We find that \tokdetok{} is less successful in improving sequence classification and ranking performance than on word-level tasks.
The \textsc{None+\ppt} setup obtains the best results in most models on the Emoji and QA datasets, suggesting the improvements seen in \tokdetok{}-based models is mostly attributable to the domain shift introduced by the Twitter pre-training corpus.\footnote{We note the complete inconsistency in both model performance and comparative model ranking present in the Emoji dataset, which calls back the systemic issue we identified in~\S\ref{sec:tasks}:
The time span over which the test set was collected contained a fundamental shift in Twitter's properties --- doubling the tweet length limit from 140 to 280 characters --- and so exhibits an unpredictable corpus incompatible with the train and dev partitions.
As a result, model performance over the dev set does not predict the test set results (a large difference on performance in this task between dev and test sets was also observed in macro-F1 scores by~\newcite{barbieri-etal-2020-tweeteval}).
In fact, under such specific data shift circumstances, it could be the case that the simpler base model is better equipped for facing longer test data, which scales generalizations made over the training and dev sets, as opposed to \tok{}-augmented models which have more levels of generalization to acquire during training and cannot anticipate the scale change.
Post-hoc inspection of specific model outputs resulted in some more concrete hypotheses for the cause of discrepancy in the major categories of confusion, for example the distributions of \say{@} presence in heart emoji
tweets and heart-eyes emoji
tweets shifted to a degree which could explain the models' growing confusion between the two, but no signals accounted for the entire difference in performance and we conclude that the main reason remains the change in sequence length distributions.}
This suggests that \tok{} may be a good learner for word-internal phenomena, picking up structural cues as to their roles within or without context, making it more useful \textbf{locally} than subword tokens' uninformed embeddings, while not being strong enough to provide a better semantic prior for \mmod{} to aggregate together with surrounding well-formed words, making its \textbf{global} utility limited.

\begin{table}
    \centering
    \small
    \begin{tabular}{lrrr}
        \toprule
        Ablation & BERT & GPT2 & RoBERTa \\
        \midrule
        Full & 37.17 & 38.27 & 41.97 \\
        No FT & $-$0.31 & $-$0.25 & $-$0.80 \\
        Wiki \ppt{} & $-$3.11 & $-$2.67 & $-$2.17 \\
        No loops & $+$0.82 & $+$0.26 & $-$1.12 \\
        All multi & $-$1.39 & $+$0.44 & $-$1.81 \\	
        \bottomrule
    \end{tabular}
    \caption{Dev set average effect of model variants compared with the \textit{All no-suff} condition: 
    \say{No-FT} --- pre-trained model used only for feature extraction;
    \say{Wiki} --- \tokdetok{} trained on Wikipedia data instead of Twitter;
    \say{No loops} --- trained without the cycle dependency loops;
    \say{All multi} --- common-suffix words also inferred using \tok{}.}
    \label{tab:ablations}
\end{table}


\subsection{Ablations}
\label{ssec:abl}
We compare the dev set results of the \textsc{All no-suff} condition on several modified versions of the model, presented in Table~\ref{tab:ablations}.
First, we find that fine-tuning \mmod{} and \tokdetok{} parameters during downstream task application is beneficial for results across base models, indicating susceptibility of \tok{}'s network to tune itself on task data and not solely on LM and the vectorization signal.
Next, and most substantially, we note the vast improvement of Twitter-trained models compared with \ppt{} performed over a Wikipedia corpus with comparable size;
even though some tasks are on edited text, the overall effect of domain change during second pre-training is apparent (and, indeed, least impactful on the NYTWIT task which features the best-edited text).
Other decisions made during the \ppt{} phase appear to be less decisive: removing the dependency loops helps performance on BERT and GPT2, but makes a large dent in the best-performing RoBERTa, indicating potential gains to be made by applying \detok{} in generative tasks not pursued within our scope;
the suffix-based dialing down of inference in pre-training helps the masked models but hurts GPT2 performance, possibly because its autoregressive application prohibits it from looking at a simply-inflected word's suffix when processing its stem, a problem not incurred in masked modeling.

\section{Related Work}
\label{sec:related}

CharBERT~\cite{ma-etal-2020-charbert} is a method which incorporates character-level encodings in \llm{}s, while re-designing the transformer stack to include a \say{character channel} distinct from the parallel token channel.
Char2Subword~\cite{aguilar2020char2subword} integrates multiple losses in a system trained to predict subword tokens from the character level, all originating in distance metrics between target and prediction.
CharacterBERT~\cite{el-boukkouri-etal-2020-characterbert} uses an ELMo-style character convolutional network to encode input into a transformer MLM.
CANINE~\cite{clark2021canine} is a method for training \llm{}s which removes the need for a subword tokenizer, by utilizing character-level representations which are pooled into inputs for the main transformer module.
ByT5~\cite{xue2021byt5} ventures yet deeper by training a T5-architecture model~\cite{raffel2020exploring} over byte representations, reducing the required embedding table size by one more order of magnitude.
Unlike \tokdetok{}, these systems all require training an \llm{} from scratch and do not support using a \ppt{} step for tuning existing large models.
In addition, with the exception of CANINE, they all resort to softmax prediction over a subword vocabulary for the generative portion of the pre-training phase, and do not offer a character-level decoder which can also be tied to the encoding component.

\section{Conclusion}
\label{sec:conclusion}

We present \tokdetok{}, an adaptive encoding-decoding system which allows large pre-trained language models to handle representation of words learned compositionally from their orthographic manifestation without resorting to hard-coded subword embeddings which fare poorly on the thick distribution tails of domain-shifted data, while still benefiting from token representations learned during the models' original pre-training phase.
We demonstrate \tokdetok{}'s efficacy across three pre-trained model architectures when trained on a corpus from a domain new to them, showing that with a relatively small amount of processing improvements can be reached on word-level tasks; with sequence-level task performance holding out in powerful pre-trained systems.
In future work, we wish to extend the scope of \tokdetok{} to non-English languages and to multilingual models, as well as to make use of the \detok{} module to assist in generative tasks.

\section*{Acknowledgments}
We thank Jon Clark, Dan Garrette, Mark Riedl, Diyi Yang, Dan Roth, and Wei Xu for their helpful comments on earlier drafts.

\bibliography{anthology,tdt}

\begin{thebibliography}{32}
\expandafter\ifx\csname natexlab\endcsname\relax\def\natexlab#1{#1}\fi

\bibitem[{{\'A}cs et~al.(2021){\'A}cs, K{\'a}d{\'a}r, and
  Kornai}]{acs2021subword}
Judit {\'A}cs, {\'A}kos K{\'a}d{\'a}r, and Andr{\'a}s Kornai. 2021.
\newblock Subword pooling makes a difference.
\newblock \emph{arXiv preprint arXiv:2102.10864}.

\bibitem[{Aguilar et~al.(2020)Aguilar, McCann, Niu, Rajani, Keskar, and
  Solorio}]{aguilar2020char2subword}
Gustavo Aguilar, Bryan McCann, Tong Niu, Nazneen Rajani, Nitish Keskar, and
  Thamar Solorio. 2020.
\newblock Char2subword: Extending the subword embedding space from pre-trained
  models using robust character compositionality.
\newblock \emph{arXiv preprint arXiv:2010.12730}.

\bibitem[{Barbieri et~al.(2020)Barbieri, Camacho-Collados, Espinosa~Anke, and
  Neves}]{barbieri-etal-2020-tweeteval}
Francesco Barbieri, Jose Camacho-Collados, Luis Espinosa~Anke, and Leonardo
  Neves. 2020.
\newblock \href {https://doi.org/10.18653/v1/2020.findings-emnlp.148}
  {{T}weet{E}val: Unified benchmark and comparative evaluation for tweet
  classification}.
\newblock In \emph{Findings of the Association for Computational Linguistics:
  EMNLP 2020}, pages 1644--1650, Online. Association for Computational
  Linguistics.

\bibitem[{Barbieri et~al.(2018)Barbieri, Camacho-Collados, Ronzano,
  Espinosa-Anke, Ballesteros, Basile, Patti, and
  Saggion}]{barbieri-etal-2018-semeval}
Francesco Barbieri, Jose Camacho-Collados, Francesco Ronzano, Luis
  Espinosa-Anke, Miguel Ballesteros, Valerio Basile, Viviana Patti, and Horacio
  Saggion. 2018.
\newblock \href {https://doi.org/10.18653/v1/S18-1003} {{S}em{E}val 2018 task
  2: Multilingual emoji prediction}.
\newblock In \emph{Proceedings of The 12th International Workshop on Semantic
  Evaluation}, pages 24--33, New Orleans, Louisiana. Association for
  Computational Linguistics.

\bibitem[{Blitzer et~al.(2007)Blitzer, Dredze, and
  Pereira}]{blitzer-etal-2007-biographies}
John Blitzer, Mark Dredze, and Fernando Pereira. 2007.
\newblock \href {https://www.aclweb.org/anthology/P07-1056} {Biographies,
  {B}ollywood, boom-boxes and blenders: Domain adaptation for sentiment
  classification}.
\newblock In \emph{Proceedings of the 45th Annual Meeting of the Association of
  Computational Linguistics}, pages 440--447, Prague, Czech Republic.
  Association for Computational Linguistics.

\bibitem[{Chen et~al.(2018)Chen, Yang, Li, Chen, and
  Wang}]{chen-etal-2018-peperomia}
Jing Chen, Dechuan Yang, Xilian Li, Wei Chen, and Tengjiao Wang. 2018.
\newblock \href {https://doi.org/10.18653/v1/S18-1067} {Peperomia at
  {S}em{E}val-2018 task 2: Vector similarity based approach for emoji
  prediction}.
\newblock In \emph{Proceedings of The 12th International Workshop on Semantic
  Evaluation}, pages 428--432, New Orleans, Louisiana. Association for
  Computational Linguistics.

\bibitem[{Clark et~al.(2021)Clark, Garrette, Turc, and
  Wieting}]{clark2021canine}
Jonathan~H Clark, Dan Garrette, Iulia Turc, and John Wieting. 2021.
\newblock Canine: Pre-training an efficient tokenization-free encoder for
  language representation.
\newblock \emph{arXiv preprint arXiv:2103.06874}.

\bibitem[{Coster et~al.(2018)Coster, van Dalen, and
  Stierman}]{coster-etal-2018-hatching}
Jo{\"e}l Coster, Reinder~Gerard van Dalen, and Nathalie Adri{\"e}nne~Jacqueline
  Stierman. 2018.
\newblock \href {https://doi.org/10.18653/v1/S18-1070} {Hatching chick at
  {S}em{E}val-2018 task 2: Multilingual emoji prediction}.
\newblock In \emph{Proceedings of The 12th International Workshop on Semantic
  Evaluation}, pages 445--448, New Orleans, Louisiana. Association for
  Computational Linguistics.

\bibitem[{Derczynski et~al.(2017)Derczynski, Nichols, van Erp, and
  Limsopatham}]{derczynski-etal-2017-results}
Leon Derczynski, Eric Nichols, Marieke van Erp, and Nut Limsopatham. 2017.
\newblock \href {https://doi.org/10.18653/v1/W17-4418} {Results of the
  {WNUT}2017 shared task on novel and emerging entity recognition}.
\newblock In \emph{Proceedings of the 3rd Workshop on Noisy User-generated
  Text}, pages 140--147, Copenhagen, Denmark. Association for Computational
  Linguistics.

\bibitem[{Devlin et~al.(2019)Devlin, Chang, Lee, and
  Toutanova}]{devlin-etal-2019-bert}
Jacob Devlin, Ming-Wei Chang, Kenton Lee, and Kristina Toutanova. 2019.
\newblock \href {https://doi.org/10.18653/v1/N19-1423} {{BERT}: Pre-training of
  deep bidirectional transformers for language understanding}.
\newblock In \emph{Proceedings of the 2019 Conference of the North {A}merican
  Chapter of the Association for Computational Linguistics: Human Language
  Technologies, Volume 1 (Long and Short Papers)}, pages 4171--4186,
  Minneapolis, Minnesota. Association for Computational Linguistics.

\bibitem[{El~Boukkouri et~al.(2020)El~Boukkouri, Ferret, Lavergne, Noji,
  Zweigenbaum, and Tsujii}]{el-boukkouri-etal-2020-characterbert}
Hicham El~Boukkouri, Olivier Ferret, Thomas Lavergne, Hiroshi Noji, Pierre
  Zweigenbaum, and Jun{'}ichi Tsujii. 2020.
\newblock \href {https://doi.org/10.18653/v1/2020.coling-main.609}
  {{C}haracter{BERT}: Reconciling {ELM}o and {BERT} for word-level
  open-vocabulary representations from characters}.
\newblock In \emph{Proceedings of the 28th International Conference on
  Computational Linguistics}, pages 6903--6915, Barcelona, Spain (Online).
  International Committee on Computational Linguistics.

\bibitem[{Kingma and Ba(2015)}]{adam}
Diederik~P Kingma and Jimmy Ba. 2015.
\newblock Adam: A method for stochastic optimization.
\newblock In \emph{ICLR}.

\bibitem[{Klein and Tsarfaty(2020)}]{klein-tsarfaty-2020-getting}
Stav Klein and Reut Tsarfaty. 2020.
\newblock \href {https://doi.org/10.18653/v1/2020.sigmorphon-1.24} {Getting the
  {\#}{\#}life out of living: How adequate are word-pieces for modelling
  complex morphology?}
\newblock In \emph{Proceedings of the 17th SIGMORPHON Workshop on Computational
  Research in Phonetics, Phonology, and Morphology}, pages 204--209, Online.
  Association for Computational Linguistics.

\bibitem[{Kudo(2018)}]{kudo-2018-subword}
Taku Kudo. 2018.
\newblock \href {https://doi.org/10.18653/v1/P18-1007} {Subword regularization:
  Improving neural network translation models with multiple subword
  candidates}.
\newblock In \emph{Proceedings of the 56th Annual Meeting of the Association
  for Computational Linguistics (Volume 1: Long Papers)}, pages 66--75,
  Melbourne, Australia. Association for Computational Linguistics.

\bibitem[{Lazaridou et~al.(2021)Lazaridou, Kuncoro, Gribovskaya, Agrawal,
  Liska, Terzi, Gimenez, d'Autume, Ruder, Yogatama
  et~al.}]{lazaridou2021pitfalls}
Angeliki Lazaridou, Adhiguna Kuncoro, Elena Gribovskaya, Devang Agrawal, Adam
  Liska, Tayfun Terzi, Mai Gimenez, Cyprien de~Masson d'Autume, Sebastian
  Ruder, Dani Yogatama, et~al. 2021.
\newblock Pitfalls of static language modelling.
\newblock \emph{arXiv preprint arXiv:2102.01951}.

\bibitem[{Liu et~al.(2019)Liu, Ott, Goyal, Du, Joshi, Chen, Levy, Lewis,
  Zettlemoyer, and Stoyanov}]{liu2019roberta}
Yinhan Liu, Myle Ott, Naman Goyal, Jingfei Du, Mandar Joshi, Danqi Chen, Omer
  Levy, Mike Lewis, Luke Zettlemoyer, and Veselin Stoyanov. 2019.
\newblock Roberta: A robustly optimized bert pretraining approach.
\newblock \emph{arXiv preprint arXiv:1907.11692}.

\bibitem[{Ma et~al.(2020{\natexlab{a}})Ma, Liu, Wang, and
  Vosoughi}]{ma-etal-2020-multi}
Weicheng Ma, Ruibo Liu, Lili Wang, and Soroush Vosoughi. 2020{\natexlab{a}}.
\newblock \href {https://doi.org/10.18653/v1/2020.emnlp-main.542}
  {Multi-resolution annotations for emoji prediction}.
\newblock In \emph{Proceedings of the 2020 Conference on Empirical Methods in
  Natural Language Processing (EMNLP)}, pages 6684--6694, Online. Association
  for Computational Linguistics.

\bibitem[{Ma et~al.(2020{\natexlab{b}})Ma, Cui, Si, Liu, Wang, and
  Hu}]{ma-etal-2020-charbert}
Wentao Ma, Yiming Cui, Chenglei Si, Ting Liu, Shijin Wang, and Guoping Hu.
  2020{\natexlab{b}}.
\newblock \href {https://doi.org/10.18653/v1/2020.coling-main.4} {{C}har{BERT}:
  Character-aware pre-trained language model}.
\newblock In \emph{Proceedings of the 28th International Conference on
  Computational Linguistics}, pages 39--50, Barcelona, Spain (Online).
  International Committee on Computational Linguistics.

\bibitem[{Nguyen et~al.(2016)Nguyen, Rosenberg, Song, Gao, Tiwary, Majumder,
  and Deng}]{nguyen2016ms}
Tri Nguyen, Mir Rosenberg, Xia Song, Jianfeng Gao, Saurabh Tiwary, Rangan
  Majumder, and Li~Deng. 2016.
\newblock Ms marco: A human generated machine reading comprehension dataset.
\newblock In \emph{CoCo@ NIPS}.

\bibitem[{Peng et~al.(2018)Peng, Thomson, and
  Smith}]{peng-etal-2018-backpropagating}
Hao Peng, Sam Thomson, and Noah~A. Smith. 2018.
\newblock \href {https://doi.org/10.18653/v1/P18-1173} {Backpropagating through
  structured argmax using a {SPIGOT}}.
\newblock In \emph{Proceedings of the 56th Annual Meeting of the Association
  for Computational Linguistics (Volume 1: Long Papers)}, pages 1863--1873,
  Melbourne, Australia. Association for Computational Linguistics.

\bibitem[{Pinter et~al.(2020)Pinter, Jacobs, and
  Bittker}]{pinter-etal-2020-nytwit}
Yuval Pinter, Cassandra~L. Jacobs, and Max Bittker. 2020.
\newblock \href {https://doi.org/10.18653/v1/2020.coling-main.572} {{NYTWIT}: A
  dataset of novel words in the {N}ew {Y}ork {T}imes}.
\newblock In \emph{Proceedings of the 28th International Conference on
  Computational Linguistics}, pages 6509--6515, Barcelona, Spain (Online).
  International Committee on Computational Linguistics.

\bibitem[{Radford et~al.(2018)Radford, Narasimhan, Salimans, and
  Sutskever}]{radford2018improving}
Alec Radford, Karthik Narasimhan, Tim Salimans, and Ilya Sutskever. 2018.
\newblock Improving language understanding by generative pre-training.
\newblock \emph{arXiv preprint}.

\bibitem[{Radford et~al.(2019)Radford, Wu, Child, Luan, Amodei, and
  Sutskever}]{radford2019language}
Alec Radford, Jeffrey Wu, Rewon Child, David Luan, Dario Amodei, and Ilya
  Sutskever. 2019.
\newblock Language models are unsupervised multitask learners.
\newblock \emph{OpenAI blog}, 1(8):9.

\bibitem[{Raffel et~al.(2020)Raffel, Shazeer, Roberts, Lee, Narang, Matena,
  Zhou, Li, and Liu}]{raffel2020exploring}
Colin Raffel, Noam Shazeer, Adam Roberts, Katherine Lee, Sharan Narang, Michael
  Matena, Yanqi Zhou, Wei Li, and Peter~J Liu. 2020.
\newblock Exploring the limits of transfer learning with a unified text-to-text
  transformer.
\newblock \emph{Journal of Machine Learning Research}, 21:1--67.

\bibitem[{Ruder(2021)}]{ruder2021lmfinetuning}
Sebastian Ruder. 2021.
\newblock {Recent Advances in Language Model Fine-tuning}.
\newblock \url{http://ruder.io/recent-advances-lm-fine-tuning}.

\bibitem[{Sennrich et~al.(2016)Sennrich, Haddow, and
  Birch}]{sennrich-etal-2016-neural}
Rico Sennrich, Barry Haddow, and Alexandra Birch. 2016.
\newblock \href {https://doi.org/10.18653/v1/P16-1162} {Neural machine
  translation of rare words with subword units}.
\newblock In \emph{Proceedings of the 54th Annual Meeting of the Association
  for Computational Linguistics (Volume 1: Long Papers)}, pages 1715--1725,
  Berlin, Germany. Association for Computational Linguistics.

\bibitem[{Strauss et~al.(2016)Strauss, Toma, Ritter, de~Marneffe, and
  Xu}]{strauss-etal-2016-results}
Benjamin Strauss, Bethany Toma, Alan Ritter, Marie-Catherine de~Marneffe, and
  Wei Xu. 2016.
\newblock \href {https://www.aclweb.org/anthology/W16-3919} {Results of the
  {WNUT}16 named entity recognition shared task}.
\newblock In \emph{Proceedings of the 2nd Workshop on Noisy User-generated Text
  ({WNUT})}, pages 138--144, Osaka, Japan. The COLING 2016 Organizing
  Committee.

\bibitem[{Sun et~al.(2019)Sun, Qiu, Xu, and Huang}]{sun2019fine}
Chi Sun, Xipeng Qiu, Yige Xu, and Xuanjing Huang. 2019.
\newblock How to fine-tune {BERT} for text classification?
\newblock In \emph{China National Conference on Chinese Computational
  Linguistics}, pages 194--206. Springer.

\bibitem[{Tjong Kim~Sang and
  De~Meulder(2003)}]{tjong-kim-sang-de-meulder-2003-introduction}
Erik~F. Tjong Kim~Sang and Fien De~Meulder. 2003.
\newblock \href {https://www.aclweb.org/anthology/W03-0419} {Introduction to
  the {C}o{NLL}-2003 shared task: Language-independent named entity
  recognition}.
\newblock In \emph{Proceedings of the Seventh Conference on Natural Language
  Learning at {HLT}-{NAACL} 2003}, pages 142--147.

\bibitem[{Vaswani et~al.(2017)Vaswani, Shazeer, Parmar, Uszkoreit, Jones,
  Gomez, Kaiser, and Polosukhin}]{vaswani2017attention}
Ashish Vaswani, Noam Shazeer, Niki Parmar, Jakob Uszkoreit, Llion Jones,
  Aidan~N Gomez, {\L}ukasz Kaiser, and Illia Polosukhin. 2017.
\newblock Attention is all you need.
\newblock \emph{Advances in Neural Information Processing Systems},
  30:5998--6008.

\bibitem[{Wolf et~al.(2020)Wolf, Debut, Sanh, Chaumond, Delangue, Moi, Cistac,
  Rault, Louf, Funtowicz, Davison, Shleifer, von Platen, Ma, Jernite, Plu, Xu,
  Le~Scao, Gugger, Drame, Lhoest, and Rush}]{wolf-etal-2020-transformers}
Thomas Wolf, Lysandre Debut, Victor Sanh, Julien Chaumond, Clement Delangue,
  Anthony Moi, Pierric Cistac, Tim Rault, Remi Louf, Morgan Funtowicz, Joe
  Davison, Sam Shleifer, Patrick von Platen, Clara Ma, Yacine Jernite, Julien
  Plu, Canwen Xu, Teven Le~Scao, Sylvain Gugger, Mariama Drame, Quentin Lhoest,
  and Alexander Rush. 2020.
\newblock \href {https://doi.org/10.18653/v1/2020.emnlp-demos.6} {Transformers:
  State-of-the-art natural language processing}.
\newblock In \emph{Proceedings of the 2020 Conference on Empirical Methods in
  Natural Language Processing: System Demonstrations}, pages 38--45, Online.
  Association for Computational Linguistics.

\bibitem[{Xue et~al.(2021)Xue, Barua, Constant, Al-Rfou, Narang, Kale, Roberts,
  and Raffel}]{xue2021byt5}
Linting Xue, Aditya Barua, Noah Constant, Rami Al-Rfou, Sharan Narang, Mihir
  Kale, Adam Roberts, and Colin Raffel. 2021.
\newblock Byt5: Towards a token-free future with pre-trained byte-to-byte
  models.
\newblock \emph{arXiv preprint arXiv:2105.13626}.

\end{thebibliography}
\bibliographystyle{acl_natbib}

\clearpage



\end{document}